\documentclass{bmvc2k}
\usepackage{floatrow}
\usepackage{enumitem}
\usepackage{xcolor}
\usepackage{booktabs}
\usepackage{colortbl}
\usepackage{amsmath}
\usepackage{amssymb}
\usepackage{amsfonts}
\usepackage[rightcaption]{sidecap}
\definecolor{mygray}{gray}{1.0}

\usepackage{float}

\title{Rich Semantics Improve Few-Shot Learning}

\addauthor{Mohamed Afham}{afhamaflal9@gmail.com}{ 1,2}
\addauthor{Salman Khan}{salman.khan@mbzuai.ac.ae}{ 2,3}
\addauthor{Muhammad Haris Khan}{muhammad.haris@mbzuai.ac.ae}{ 2}
\addauthor{Muzammal Naseer}{muzammal.naseer@mbzuai.ac.ae}{ 3,2}
\addauthor{Fahad Shahbaz Khan}{fahad.khan@mbzuai.ac.ae}{ 2,4}

\addinstitution{
 Department of Electronic and Telecommunication Engineering,\\
 University of Moratuwa,\\
 Sri Lanka
}
\addinstitution{
 Mohamed Bin Zayed University of AI,\\
 UAE
}
\addinstitution{
 Australian National University,\\
 AU
}

\addinstitution{
 Link\"oping University,\\
 Sweden
}

\runninghead{Afham et al.}{Rich Semantics Improve Few-Shot Learning}

\def\eg{\emph{e.g}\bmvaOneDot}

\def\etal{\emph{et al}\bmvaOneDot}

\begin{document}

\maketitle

\begin{abstract}
\noindent
Human learning benefits from multi-modal inputs that often appear as rich semantics (e.g., description of an object's attributes while learning about it). This enables us to learn generalizable concepts from very limited visual examples. However, current few-shot learning (FSL) methods use numerical class labels to denote object classes which do not provide rich semantic meanings about the learned concepts. In this work, we show that by using  `class-level' language descriptions, that can be acquired with minimal annotation cost, we can improve the FSL performance. Given a support set and queries, our main idea is to create a bottleneck visual feature (hybrid prototype) which is then used to generate language descriptions of the classes as an auxiliary task during training. We develop a Transformer based forward and backward encoding mechanism to relate visual and semantic tokens that can encode intricate relationships between the two modalities. Forcing the prototypes to retain semantic information about class description acts as a regularizer on the visual features, improving their generalization to novel classes at inference. Further, this strategy imposes a human prior on the learned representations, ensuring that the model is faithfully relating visual and semantic concepts, thereby improving model interpretability. Our experiments on four datasets and ablations show the benefit of effectively modeling rich semantics for FSL. Code is available at: {\small \url{https://github.com/MohamedAfham/RS_FSL}}.
\end{abstract}
\section{Introduction}
Traditional classification models use class labels for supervision, expressed in a numerical form or as one-hot encoded vectors \cite{khan2018guide}. However, humans do not solely rely on such numerical class-labels to acquire learning. Instead, humans learn by communicating through natural language, which is grounded in a complex structure consisting of semantic attributes, relationships and abstract representations. Psychologists and cognitive scientists have argued natural language descriptions to be a central element of human learning \cite{smith2003learning,chin2017contrastive,miller2019explanation}. The depiction of semantic class labels with numerical IDs leads to a \emph{semantic gap} between the class semantic representation and the learned visual features.
\begin{SCfigure}[50]
    \caption{\small In FSL setting, where we require generalizability to novel classes with limited samples, modeling semantic attributes of classes can help disambiguate confusing classes. We suggest that the numerical class labels traditionally used in FSL are inadequate to represent diverse semantic attributes of an object class, which can be modeled via low-cost class-level language descriptions (colored boxes). Our approach effectively utilizes language information to learn both highly discriminative and transferable visual representations that help to avoid errors in ambiguous cases (e.g., visually similar fine-grained classes).}
    \includegraphics[width=0.4 \linewidth]{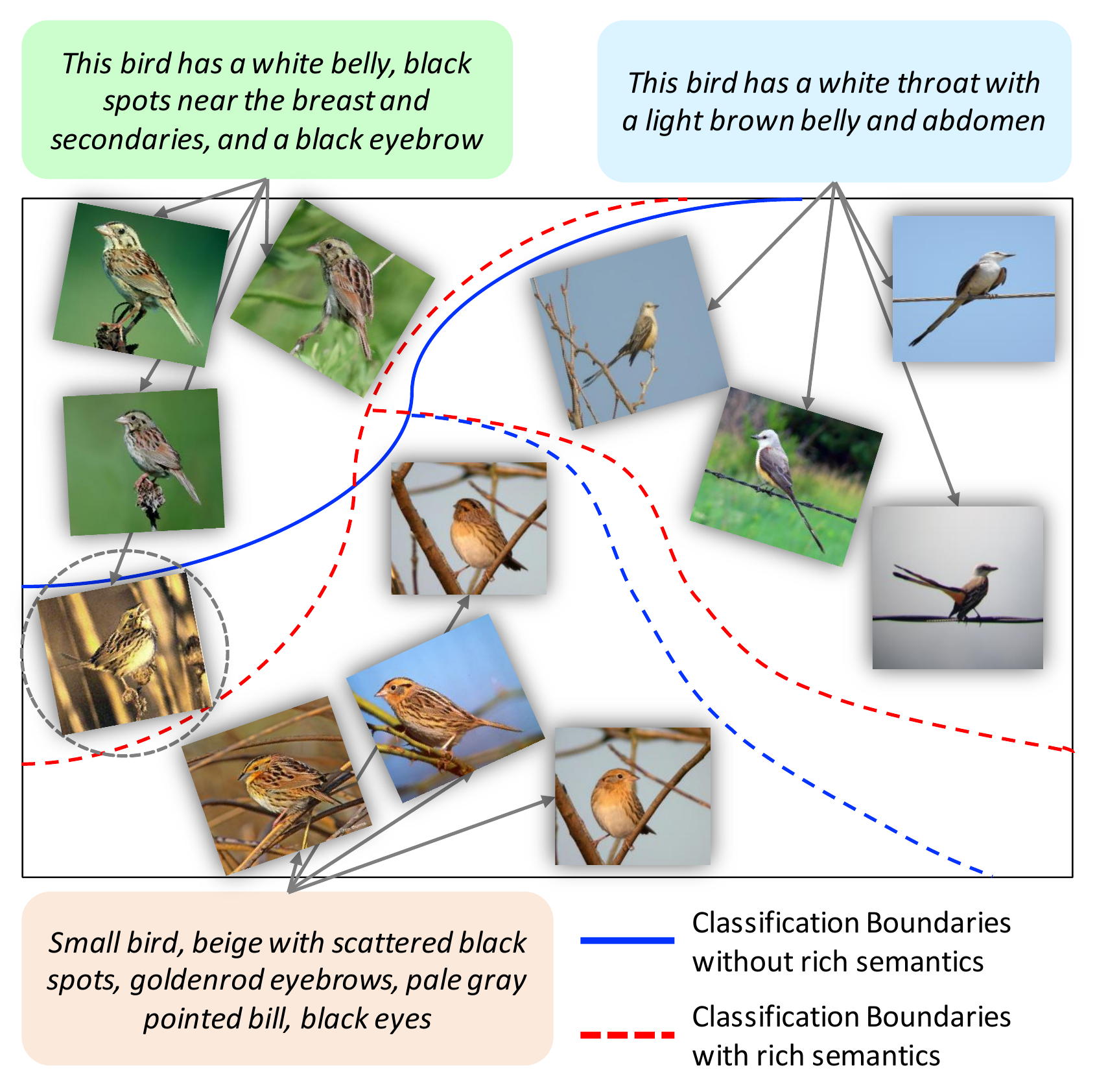}
    \label{fig:abstract}
\end{SCfigure}

We consider a few-shot learning setting where language descriptions for the seen (\emph{base}) classes are available during training but not for the novel (\emph{few-shot}) class that appear during inference.  Remarkably, studying the potential of language descriptions has particular relevance to FSL, where a model must learn to generalize from few-samples and several categories can only be discriminated with subtle attribute-based differences (Fig.~\ref{fig:abstract}). We hypothesize that by predicting natural language descriptions as an auxiliary task during training, the model can learn useful representations that help transfer better to novel tasks during the inference stage. This helps the representations to explicitly model the shared semantics between the few-shot samples so that the class descriptions can be successfully generated. 

The language description task while learning to classify images forces the model to attain following desirable attributes: (a) model high-level compositional patterns occurring in the visual data e.g., attributes in fine-grained bird classes; (b) avoid over-fitting on a given FSL task by imposing a regularizer demanding natural description from the class prototype; and (c) provide intuitive explanation for the learned class concepts e.g., the description of an object type, attributes, function and affordance in a human interpretable form. Importantly, the error feedback obtained from such a supervised task (natural language description) can help align a model with the `\emph{human prior}’. 
Our RS-FSL approach is generic in nature and can be plugged into existing baselines models or with other multi-task objectives (e.g., equivariant self-supervised learning losses). Although we discard the language description module at inference, it is useful as a debugging tool to understand the model's behaviour in case of wrong predictions (\eg, highlighting which attributes were mistaken or ambiguous).\\

\noindent \textbf{Contributions.} Our objective induces a generative task of natural language description for few-shot classes that forces the model to learn correlations between same class samples such that consistent class descriptions can be generated. The limited set of class-specific samples acts as a bottleneck that encourages extraction of shared semantics. We then introduce a novel transformer decoding approach for few-shot class description that relates the hybrid prototypes (obtained using the collective support and query image features) with the corresponding descriptions in both forward and backward directions. Our design allows modeling long-range dependencies between the sequence elements compared to previous recurrent architectures. Finally, our experiments on four datasets show consistent improvements across the FSL tasks. We extensively ablate our approach and analyze its different variants. The proposed transformer decoder acts as a plug-and-play module and shows gains across popular FSL baselines namely ProtoNet \cite{snell2017prototypical}, RFS \cite{RFS} and Meta-baseline \cite{chen2020new}. 
\section{Related Work}
Only a few-methods explore the potential of rich semantic descriptions in the context of FSL. A predominant approach has been the incorporation of unsupervised word embeddings or a set of manual attributes to represent class semantics. For example, \cite{chen2019multi} uses semantic embeddings of the labels or attributes to guide the latent representation of an auto-encoder as a regularizer mechanism.  Xing \etal \cite{xing2019adaptive} dynamically learn the class prototypes as a convex combination of visual and semantic label embeddings (based on GloVe \cite{pennington2014glove}). However, these embeddings require manual labeling (in case of attributes) or remain noisy if acquired via unsupervised learning. Additionally, representing rich semantics in a single vector remains less flexible to encode the complex semantics. In contrast, our approach flexibly learns the semantic representations with class-level language descriptions to improve upon the noisy unsupervised word embeddings. Schwartz \etal \cite{schwartz2019baby} extended \cite{xing2019adaptive}  to exploit various semantics (category labels, text descriptions and manual attributes) in a joint framework. However, they use language descriptions as inputs rather than an extra supervision signal to train the visual backbone. Thus, these methods \cite{xing2019adaptive,schwartz2019baby} require attribute information or descriptions for novel classes during inference which can be hard to acquire for few-shot classes.

Image-level captions have been used in \cite{zareian2020open,desai2020virtex} to align visual and semantic spaces with a multi-modal transformer model \cite{khan2021transformers}. This can help learning from a limited set of base classes and scales to unseen classes \cite{zareian2020open}. \cite{Donahue2011annotator} models annotator rationale as a spatial attention and the relevant attributes for a given input image.  However, unlike our work, these methods do not study the FSL problem where image-level captions can cause overfitting. Furthermore, they require image-specific captions and rationales (not just “what” but also “why”) which can be costly, even for a small number of base classes. Since acquiring high-quality explanations \cite{mu2020shaping, andreas2017learning} from experts can be expensive, efforts have been made to reduce the manual cost needed to acquire such annotations. To this end, ALICE model \cite{liang2020alice} acquires contrastive natural language descriptions from human annotators about the most informative pairs of object classes identified via active learning. In contrast, our approach only requires class-level descriptions that are easy to acquire compared to image level semantic annotations.

Andreas \etal \cite{andreas2017learning} use the language descriptions during the pertaining stage in FSL to learn natural task structure. Once the model is pretrained to match images to natural descriptions, it can be used to learn new concepts by aligning natural descriptions with the images at inference. In contrast to inference stage alignment, LSL \cite{mu2020shaping} introduced a GRU branch with language supervision to enrich the backbone features and discards the branch during inference. However, decoding mechanism in \cite{mu2020shaping,andreas2017learning} does not explicitly encode both forward and backward relations in the language, suffers in encoding long-term relationships and cannot relate multiple class-level descriptions with the visual class prototypes. 


\begin{figure}[t]
    \centering
    \includegraphics[width=1 \linewidth]{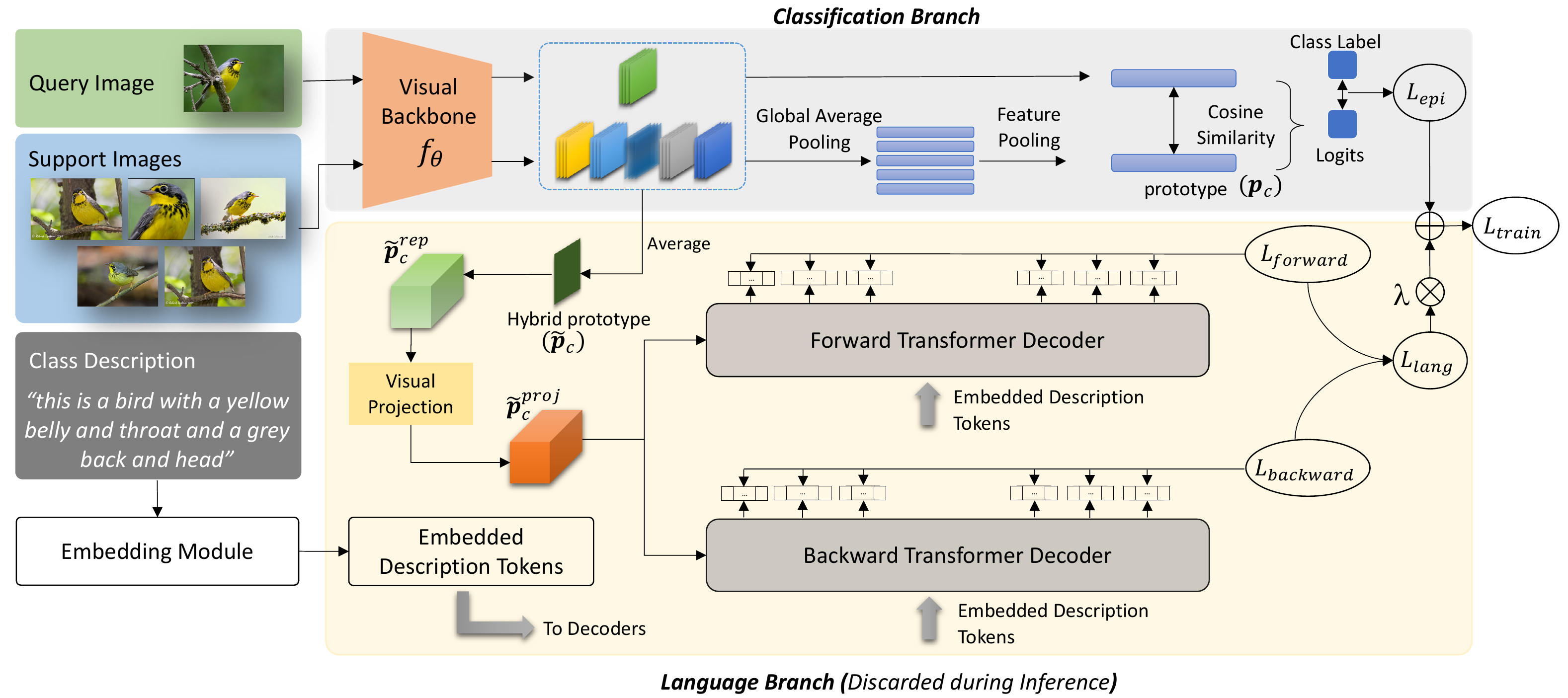}\vspace{-1em}
    \caption{Overall architecture of the RS-FSL. Fundamentally, it consists of a visual backbone followed by a prototypical network to compute the classification loss (denoted by classification branch). We aim at encoding visual-semantic relationships that are in turn harnessed for enriching the visual features. To this end, we propose generating class-level language descriptions constrained on a hybrid prototype via developing a transformer based forward and backward decoding mechanism (denoted as language branch). Our method jointly trains the classification and language losses.} 
    \label{fig:overall_architecture}
\end{figure}
\section{Proposed Method}
\subsection{Preliminaries}
\label{subsection:Preliminaries}

\noindent \textbf{Problem Settings.} In the standard \textit{few-shot image classification} setting, we have access to a labelled dataset of base classes $\mathcal{C}_{base}$ with enough number of examples in each class, and the aim is to learn concepts in novel classes $\mathcal{C}_{novel}$ with few examples, given $\mathcal{C}_{base} \cap \mathcal{C}_{novel}= \emptyset$. 

\noindent \textbf{Pretraining.}  Following the recent works in FSL \cite{chen2020new,simpleshot}, we pretrain the \textit{visual backbone}, parameterized by $\theta$, on $\mathcal{C}_{base}$ in a supervised manner. We assume access to a dataset of image ($x$) and label ($y\in \mathcal{C}_{base}$) pairs: $\mathcal{D} = \{x_i,y_i\}_{i=1}^{M}$. The \textit{visual backbone} maps the input image $x$ to a feature embedding space $\mathbb{R}^{d}$ by an embedding function $f_{\theta}:x\rightarrow v$. In turn, the linear classifier $f_{\Theta}:v\rightarrow p$, parameterized by $\Theta$, maps the features generated by $f_{\theta}$ to the label space $\mathbb{R}^{L}$ where $L$ denotes the number of classes in $\mathcal{C}_{base}$. We optimize both $\theta$ and $\Theta$ by minimizing the standard cross-entropy loss: \\

\vspace{-2em}
\begin{equation}
\mathcal{L}_{pre}(p,y) = -\log\frac{\exp({p_y})}{\sum_j \exp({p_j})} 
\end{equation}

\noindent \textbf{Episodic Training.} After pretraining the visual backbone, we adopt the episodic training paradigm which has shown effectiveness for FSL. It simulates few-shot scenario faced at test time via constructing episodes by sampling small number of few-shot classes from a large labelled collection of classes $\mathcal{C}_{base}$. Specifically, each episode is created by  sampling $N$ classes from the $\mathcal{C}_{base}$ forming a support class set $\mathcal{C}_{supp} \subset \mathcal{C}_{base}$. Then two different example sets are sampled from these classes. The first is a \textit{support-set} $S_{e} = \{(s_{i},y_{i})\}_{i=1}^{N \times K}$ comprising \textit{K} examples from each of $N$ classes, and the second is a \textit{query-set} $Q_{e} = \{(q_{j},y_{j})\}_{j=1}^{Q}$ containing \textit{Q} examples from the same \textit{N} classes. The episodic training for few-shot classification boils down to minimizing, for each episode $e$, the loss of prediction on the examples in the query-set $(q_{j},y_{j}) \in Q_{e}$, given the support set $S_{e}$: 
\vspace{-0.2cm}
\begin{equation}
    \mathcal{L}_{epi} = \mathbb{E}_{(S_{e},Q_{e})}\sum_{j=1}^{Q}\log {P}_{\theta}(y_{j}|q_{j},S_{e}). \label{Eq:episodic_loss}
\end{equation}

\noindent \textbf{Prototypical Networks.} We develop proposed method on a popular metric-based meta-learning method named Prototypical network \cite{snell2017prototypical}, owing to its simplicity and effectiveness. However, ours is a plug-and-play training module which can work seamlessly with other FSL methods (as demonstrated in Sec.~\ref{subsection:ablation_study}). Prototypical networks leverage the support set to compute a centroid (a.k.a \emph{prototype}) for each class in a given episode, and query examples are classified based on distance to each prototype. The model is a convolutional neural network with parameters $\theta$, that learns a $d$-dimensional space where examples from the same class are clustered together and those of different classes are far apart. Formally, for each episode $e$, a prototype $\textbf{p}_{c}$ corresponding to class $c \in \mathcal{C}_{base}$ is computed by averaging the embeddings of all support samples belonging to class $c$:

\begin{equation}
    \textbf{p}_{c}=\frac{1}{|S_{e}^{c}|}\sum_{(s_{i},y_{i}) \in S_{e}^{c}} f_\theta (x), \label{Eq:proto_compute}
\end{equation}

\noindent where $f_\theta$ is the pretrained visual backbone, and $S_{e}^{c}$ is the subset of support belonging to class $c$. The model generates a distribution over $N$ classes in an episode after applying softmax over cosine similarities between the embedding of the query $q_{j}$ and the prototypes $\textbf{p}_{c}$:
\begin{equation}
    P_{\theta}(y = c| q_{j}, S_{e}) = \frac{ \exp(\tau.\langle f_{\theta}(q_{j}), \textbf{p}_{c}\rangle)}{\sum_{k}\exp(\tau.\langle f_{\theta}(q_{j}), \textbf{p}_{k} \rangle)}, 
    \label{Eq:proto_classify}
\end{equation}
\noindent where $\langle.,.\rangle$ is the cosine similarity, $k \in \mathcal{C}_{supp}$, and $\tau$ is the learnable parameter to scale the cosine similarity for computing logits \cite{chen2020new}. The model is trained by minimizing Eq.~\ref{Eq:episodic_loss}.
In the following section, we propose capturing rich and shared class-level semantics for FSL tasks via predicting natural language descriptions.

\subsection{Capturing Rich Semantics for FSL}
\label{subsection:RS_FSL}
We show that by leveraging class-level semantic descriptions, the performance of FSL tasks can be improved. To this end, we create a bottleneck visual feature (termed hybrid prototype) to generate the language descriptions of classes as an auxiliary task. We introduce a language description branch featuring a Transformer based forward and backward decoding mechanism to connect hybrid prototypes with the corresponding descriptions both in forward and backward directions. This enforces the model to capture correlations between the same class examples so as to produce consistent class level descriptions.The hybrid prototype that is obtained using the support and query visual features facilitates the extraction of shared semantics. Furthermore, our language branch allows modelling long-range dependencies between the sequence of token vectors compared to prior recurrent architectures. We elaborate these components below. 

\noindent \textbf{Mapping visual features to language description.}
For each class $c \in \mathcal{C}_{base}$, we assume to have $d_c$ \textit{class-level} language descriptions $W_{c}=\{w_1,w_2,...,w_{d_c}\}$. Each $w_i = (w_{i,1},w_{i,2},....w_{i,T_i}): i \in \{1, d_c\}$ is a language description of variable length $T_i$ tokens where $w_{i,1}=\langle s \rangle$ represents the start of the sentence token and $w_{i,T_i}=\langle /s \rangle$ denotes the end of the sentence token. Let $\tilde{\textbf{p}}_{c}$ be the hybrid prototype formed after averaging the embeddings of support  examples $S_{e}^{c}$ and query examples $Q_{e}^{c}$ of class $c$, as follows:
\vspace{-0.5em}
\begin{equation}
    \tilde{\textbf{p}}_{c}=\frac{1}{|S_{e}^{c}| + |Q_{e}^{c}|}\sum_{x\in (S_{e}^{c} \cup \ Q_{e}^{c} )} f_\theta (x). \label{Eq:new_proto}
\end{equation}
Notably, the hybrid prototype $\tilde{\textbf{p}}_{c}$ contrasts with $\textbf{p}_{c}$ that only averages the support features. Our language module, associates the hybrid prototype $\tilde{\textbf{p}}_{c}$ with the corresponding descriptions $W_{c}$ in both forward and backward directions. Specifically, the proposed transformer decoding function $g_\phi$, parameterized by $\phi$, takes the hybrid prototype $\tilde{\textbf{p}}_{c}$ and predicts class semantic descriptions $\tilde{W_{c}}$. It comprises a forward and a backward model which allows it to generate each description $\tilde{w_i}$ token-by-token from left-to-right and right-to-left, respectively, and uses the following language loss for training:
\begin{align}
\mathcal{L}_{lang}(\theta,\phi) & = \frac{1}{2} \Big(\sum_{i=1}^N \sum_{t=2}^{T_i} -\log g_\phi \left(\tilde{w}_{i,t} = w_{i,t} \ | \ \tilde{\textbf{p}}_{c}\right)  +  \sum_{i=1}^N \sum_{t=T_i-1}^{1} -\log g_\phi \left(\tilde{w}_{i,t} = w_{i,t} \ | \ \tilde{\textbf{p}}_{c} \right)\Big). \notag
\end{align}

\begin{figure}
\floatbox[{\capbeside\thisfloatsetup{capbesideposition={right,top},capbesidewidth=6cm}}]{figure}[\FBwidth]
{\caption{Transformer decoder architecture (bottom box) for generating class-level natural language descriptions based on multi-head attention (top box). We replicate the same architecture for both forward and backward decoding mechanisms. After the last Transformer layer, we apply a linear layer to get output un-normalized log probabilities over the token vocabulary.}\label{fig:transformer_decoder}}
{\includegraphics[width=1 \linewidth]{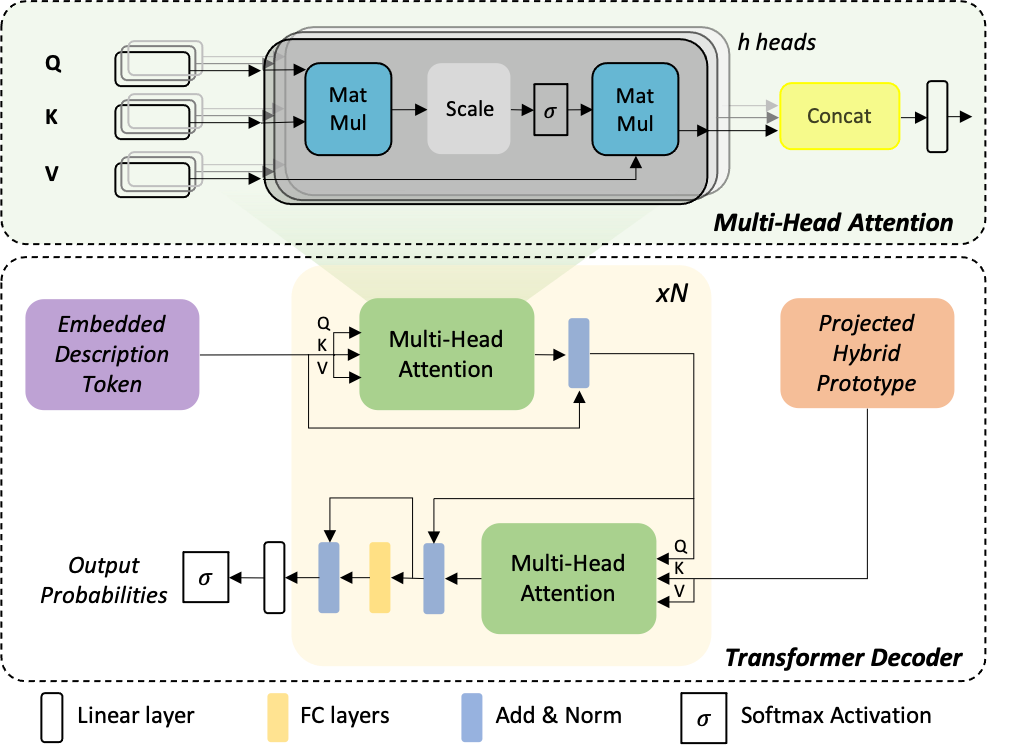}}
\end{figure}
We jointly train the (pretrained) visual backbone $f_{\theta}$ and the language description branch $g_{\phi}$, in an episodic manner, after combining both the classification and language losses:
\begin{equation}
    \mathcal{L}_{train}= \mathcal{L}_{epi} + \lambda \mathcal{L}_{lang}, \label{Eq:train_loss}
\end{equation}
\noindent where $\lambda$ balances the contribution of $\mathcal{L}_{lang}$ towards the joint loss $\mathcal{L}_{train}$. Our overall objective provides a learning signal that facilitates aligning the model with the human semantic prior of which representations are more transferable than others. We discard the language description branch after the training loop during inference. However, it can be beneficial towards understanding the model behaviour for incorrect predictions \eg finding which attributes were mistaken during inference.

\noindent \textbf{Transformer Decoder Architecture.} Inspired by recent advances in language modelling, we propose to use Transformers \cite{transformer}, for decoding class-level descriptions in both forward and backward directions (Fig.~\ref{fig:transformer_decoder}). Transformers feature multi-head self-attention and not only can propagate the contextual information over sequence of description tokens but have the expressive capability to relate the hybrid prototype to semantic tokens in class-level descriptions. In a training episode, each image is accompanied with a corresponding class description. The hybrid prototype $\tilde{\textbf{p}}_{c}$ of a given class is replicated to match the number of descriptions available in the episode and denoted by $\tilde{\textbf{p}}_{c}^{rep}$. To reduce the complexity of the resulting feature tensor, it is projected through a linear layer. The projected hybrid prototype $\tilde{\textbf{p}}_{c}^{proj}$ is then fed to the decoder module. 


We embed the class-level descriptions using the Embedding Module (Fig.~\ref{fig:overall_architecture}) which is initialised with pretrained GloVe~\cite{pennington2014glove}. The result is a set of embedded description tokens, which are fed to both forward and backward transformer decoders. The decoder first performs a multi-head self-attention over description token vectors and then applies multi-head attention between the projected hybrid prototype and descriptive token vectors. In each multi-head attention block, the inputs are transformed to query ($\textbf{Q}$), key ($\textbf{K}$) and value ($\textbf{V}$) triplets using a set of transformation matrices.  Attention mechanism is similar to \cite{transformer} where the future elements of the description are masked to perform masked multi-head attention (see architecture in Fig.~\ref{fig:transformer_decoder}). It then applies a two-layer fully connected network to each vector. All these operations are followed by dropout, enclosed in a residual connection, and followed by layer normalization. After passing through transformer layers, we apply a linear layer which is common to both forward and backward decoders of each vector to produce un-normalized log probabilities over the token vectors. Following recent works \cite{desai2020virtex}, our transformer employs GELU activation \cite{DBLP:journals/corr/HendrycksG16} instead of ReLU. 

\section{Experiments} \label{sec:experiments}
\noindent \textbf{Datasets.} 
\textbf{CU-Birds} \cite{WelinderEtal2010} is an image dataset with 200 different birds species each having 40-60 images. Following \cite{Su2020When}, we split the available classes into 100 for training, 50 for validation and 50 for testing.
\textbf{VGG-Flowers} is a fine-grained classification dataset comprising 102 flowers categories. Following \cite{Su2020When}, we split the dataset into 51 for training, 26 for validation and 25 for test classes. For both CUB and VGG-Flowers datasets, we acquire natural language descriptions for the images from \cite{reed2016learning} which provides 10 captions per image. Since our method leverages class-level descriptions, we sample the required number of descriptions from the (available) captions of the images belonging to each class, but those are consistently used for all the class images.
\textbf{miniImageNet} \cite{NIPS2016_90e13578} is a popular dataset for few-shot classification tasks. It consists of 100 image classes extracted from the original ImageNet dataset \cite{imagenet_cvpr09}. Each class contains 600 images of size $84\times84$. We follow the splitting protocol proposed by \cite{NIPS2016_90e13578}, and use 64 classes for training, 16 for validation, and 20 for testing. Since, class level descriptions for this dataset are unavailable, we manually gathered them from the web. Some representative examples of class-level descriptions for miniImageNet  are shown in the supplementary material. 
\textbf{ShapeWorld} is a synthetic multi-modal dataset proposed by \cite{Kuhnle2017ShapeWorldA}. It consists of 9000, 1000, and 4000 few-shot tasks for training, validation and testing, respectively. Each task has a single support set of $K = 4$ images that are representing a visual concept with an associated natural language description, which we consider as the class-level descriptions. Each concept describes a spatial relation between two objects, and each object is optionally qualified by color and/or shape, with 2-3 distractor shapes around. The task is to predict whether a query image belongs to the concept or not.
                 
\noindent \textbf{Implementation Details.}
For fair comparisons with prior works,
we deploy the following CNN architectures as visual backbones: 4-layer convolutional architecture proposed in \cite{snell2017prototypical} for CUB, ResNet-12 for miniImageNet \cite{RFS, chen2020new, metaoptnet}, and ResNet-18 \cite{Su2020When} for VGG-Flowers. For evaluation on all datasets, we use the challenging FSL setting of 5-way 1-shot  and report accuracy averaged across few-shot tasks along with 95\% confidence interval. During \textit{pretraining} stage, we use SGD optimizer with an initial learning rate of 0.05, momentum of 0.9, and weight decay of 0.0005. We train the model for 100 epochs with a batch size of 64 and the learning rate decays twice by a factor of 0.1 at 60 and 80 epochs. 
Both forward and backward decoders are configured with a hidden layer size of 768, 12 attention heads, a feed-forward dimension of 3072 and with a 0.1 dropout probability.
We use Adam optimizer with a constant learning rate of 0.0005 throughout and train the models for 600 epochs. Standard data augmentation e.g., random crop, color jittering and random horizontal flipping are applied during the meta-training stage. We fix $\lambda = 5$ in all experiments. $\tau$ is initialized as 1 for experiments with CUB and VGG-Flowers while for miniImageNet experiments it's initialized as 10. We use 2 layers of transformer decoders based on validation and study the effect of different layers in Fig.~\ref{fig:ablation_varying_descriptions}. We use 20 \textit{class-level} descriptions for both CUB and VGG-Flowers while for miniImageNet we use all 5 descriptions available per class. During inference, we discard the language description branch and rely on the visual backbone to perform few-shot classification. To be consistent with previous works \cite{Chen2019ACL, Su2020When}, we sample 600 few-shot tasks from the set of novel classes.
For ShapeWorld dataset, following \cite{mu2020shaping} we train for 50 epochs with a constant learning rate of 0.00005 with Adam optimizer. During training, we use $\lambda = 20$ and similar transformer decoder architecture parameters as the experiments in other datasets.  
\vspace{-0.4cm}
\subsection{Comparison with state-of-the-art}\label{sec:results}
\noindent We compare the performance of our method with eleven existing top performing approaches on CUB dataset in Tab.~\ref{table:SOTA_cub_mini} (right). Our method delivers a significant improvement of $6.36\%$ over a strong baseline method \cite{chen2020new}.
Tab.~\ref{table:SOTA_cub_mini} (left) reports experimental results on miniImageNet. RS-FSL provides an improvement of 2.16\% over the competitive baseline \cite{chen2020new}. Compared to prior works, our method attains a higher accuracy of 65.33\% and demonstrates the best performance. Experimental results for ShapeWorld dataset are reported in Tab.~\ref{table:SOTA_VGG_ShapeWorld}(b). RS-FSL outperforms the existing best performing method LSL \cite{mu2020shaping} by a margin of $1.15\%$. Further, the results for VGG-Flowers dataset are shown in Tab.~\ref{table:SOTA_VGG_ShapeWorld}(a). RS-FSL performs favorably against all competing methods and achieves the best accuracy of 75.33\%.
\begin{table}[t]
\footnotesize
\centering
\begin{tabular}{c c c|c c c}
 \toprule
 \multicolumn{3}{c|}{\textbf{miniImageNet}} & \multicolumn{3}{c}{\textbf{CUB}}  \\ [0.5ex]
 \midrule
 \rowcolor{mygray} \textbf{Method} & \textbf{Backbone} & \textbf{Accuracy} & \textbf{Method} & \textbf{Backbone} & \textbf{Accuracy} \\ [0.5ex] 
 \midrule
 ProtoNet \cite{snell2017prototypical} & Conv-4 & 55.50$\pm$0.70 & MatchingNet \cite{NIPS2016_90e13578} & Conv-4 & 60.52$\pm$0.88 \\
 Matching Net \cite{NIPS2016_90e13578}& Conv-4 & 43.56$\pm$0.78 & MAML \cite{pmlr-v70-finn17a} & Conv-4 & 54.73$\pm$0.97 \\ 
 MAML\cite{pmlr-v70-finn17a} & Conv-4 & 48.70$\pm$1.84 & ProtoNet \cite{snell2017prototypical} & Conv-4 & 50.46$\pm$0.88 \\ 
 Chen \etal\cite{Chen2019ACL} & Conv-4 & 48.24$\pm$0.75 & RFS \cite{RFS} & Conv-4 & 41.47$\pm$0.72 \\ 
 Relation Net\cite{Sung2018LearningTC} & Conv-4 & 50.44$\pm$0.82 & RelationNet \cite{Sung2018LearningTC} & Conv-4 & 62.34$\pm$0.94\\ 
 TADAM \cite{TADAM}& ResNet-12 & 58.50$\pm$0.30 & L3 \cite{andreas2017learning} & Conv-4 & 53.96$\pm$1.06 \\ 
 MetaOptNet \cite{metaoptnet} & ResNet-12 & 62.64$\pm$0.61 & LSL \cite{mu2020shaping} & Conv-4 & 61.24$\pm$0.96 \\ 
 Boosting \cite{boosting}& WRN-28-10 & 63.77$\pm$0.45 & Chen \etal \cite{Chen2019ACL} & Conv-4 & 60.53$\pm$0.83\\ 
 RFS-Simple \cite{RFS} & ResNet-12 & 62.02$\pm$0.63 & DN4-DA \cite{DNA} & Conv-4 & 53.15$\pm$0.84 \\
 RFS-Distill \cite{RFS} & ResNet-12 & 64.82$\pm$0.60 & HP \cite{Khrulkov2020HyperbolicIE} & Conv-4 & 64.02$\pm$0.24 \\
Meta-Baseline \cite{chen2020new} & ResNet-12 & 63.17$\pm$0.23 & Meta-Baseline \cite{chen2020new}  & Conv-4 & 59.30$\pm$0.86 \\ 
 \midrule
 RS-FSL  & ResNet-12 & \textbf{65.33$\pm$0.83} & RS-FSL  & Conv-4 & \textbf{65.66$\pm$0.90} \\
 \bottomrule
\end{tabular}
\vspace{-1em}
\caption{Comparison with prior works on CUB and miniImageNet. Our method, RS-FSL, allows exploiting semantic information during training only.
}
\label{table:SOTA_cub_mini}
\end{table}
Overall, RS-FSL consistently show gains on all four datasets. We note the improvement is more significant in CUB as compared to others since the class samples conform better with the language descriptions as compared to \eg, Flowers dataset. Thus our class-level descriptions show more effectiveness there. The increment on miniImageNet is relatively less pronounced due to the limited class descriptions obtained manually by us (only 5 per class). We also display some qualitative examples in supplementary material which show that RS-FSL encourages the model to focus on semantically relevant visual regions.

\begin{table}[h]
    \subtable[][Performance on VGG-Flowers]{
        \centering
        \resizebox{.3\textwidth}{!}{
        \begin{tabular}{c c} 
          \toprule
         \rowcolor{mygray} \textbf{Method} & \textbf{Accuracy} \\ [0.5ex] 
         \hline
         ProtoNet \cite{snell2017prototypical}  & 72.38$\pm$0.98 \\ 
         Matching Net \cite{NIPS2016_90e13578}& 73.51$\pm$0.94 \\ 
         MAML\cite{pmlr-v70-finn17a} &  65.46$\pm$1.05 \\ 
         Chen \etal\cite{Chen2019ACL} & 74.09$\pm$0.84 \\
         Relation Net\cite{Sung2018LearningTC} & 55.59$\pm$1.09 \\
         Meta-Baseline \cite{chen2020new} &  73.35$\pm$0.98 \\ 
         \hline
         RS-FSL  &  \textbf{75.33$\pm$0.96} \\
         \bottomrule
        \end{tabular}}
    }
    \subtable[][Performance on ShapeWorld]{
        \centering
        \resizebox{.3\textwidth}{!}{
        \begin{tabular}{c c} 
         \toprule
         \rowcolor{mygray} \textbf{Method} & \textbf{Accuracy} \\ [0.5ex] 
         \hline
         ProtoNet \cite{snell2017prototypical}$^\dagger$  & 50.91$\pm$1.10 \\ 
         L3 \cite{andreas2017learning}$^\dagger$& 62.28$\pm$1.09 \\ 
         LSL \cite{mu2020shaping} &  63.25$\pm$1.06 \\ 
         \hline
         RS-FSL & \textbf{64.40$\pm$0.99} \\
         \bottomrule
        \end{tabular}
        }
    }
    \vspace{-1em}
    \caption{Comparison on VGG-Flowers and ShapeWorld. $^\dagger$ Results reported in \cite{mu2020shaping}. }
    \label{table:SOTA_VGG_ShapeWorld}
\end{table}

\subsection{Analysis and Ablation Study}
\label{subsection:ablation_study}
We perform all ablation experiments on CUB dataset with a Conv-4 visual backbone architecture following ProtoNet \cite{snell2017prototypical} baseline.

\noindent \textbf{The effect of different baselines and word embeddings.} To demonstrate the generalizability of our approach, we show its improvements across three popular baseline FSL approaches both with and without using language prediction during training (see Tab.~\ref{table:ablation_different_baselines}). We note that the proposed language prediction mechanism constrained on a bottleneck visual feature (hybrid prototype) consistently improves the performance under all three baseline methods: ProtoNet \cite{snell2017prototypical}, RFS \cite{RFS}, and Meta-Baseline~\cite{chen2020new}. Tab.~\ref{table:ablation_different_embeddings} reports the impact on performance when using three different word embeddings to represent the words: Word2Vec \cite{word2vec} , GloVE \cite{pennington2014glove}, and fastText \cite{fasttext}. Our method retains similar accuracies under both Word2Vec and GloVe, however, it performs slightly inferior when deploying fastText. This reveals that RS-FSL is robust to the choice of word embeddings and favorable gains are obtained over the baseline method regardless of the embedding type used.


\begin{table}[h]
    \vspace{-0.5em}
        \caption{(a) Performance of different baselines both with and without our rich semantic (RS) modeling and (b) performance when using three different word embeddings.}
    \label{table:ablation_embeddings_baseline}
    \subtable[][Effect of different baselines]{
        \centering
        \resizebox{.48\textwidth}{!}{
        \begin{tabular}{c c c c} 
         \toprule
        \rowcolor{mygray} \textbf{Baseline} & \textbf{Backbone} & \begin{tabular}{@{}c@{}} \textbf{Without RS}\end{tabular}
         & \begin{tabular}{@{}c@{}}\textbf{With RS}\end{tabular} \\
         \midrule
         ProtoNet \cite{snell2017prototypical}  & Conv-4 & 57.97$\pm$0.96 & 63.86±0.91 \\ 
         RFS \cite{RFS} & Conv-4 & 44.93$\pm$0.76 & 46.84±0.86 \\
         Meta-Baseline \cite{chen2020new} & Conv-4 & 59.30±0.86 & \textbf{65.66±0.90} \\
         \bottomrule
        \end{tabular}}
        \label{table:ablation_different_baselines}
    }
    \subtable[][Impact of Different Word embeddings]{
        \centering
        \resizebox{.45\textwidth}{!}{
        \begin{tabular}{c c c c} 
         \toprule
        \rowcolor{mygray} \textbf{Word Embedding} & \textbf{Backbone} & \textbf{Accuracy} & \begin{tabular}{@{}c@{}}\textbf{$\%$ gain} \\ \textbf{over baseline}\end{tabular}\\ [0.5ex] 
         \midrule
         Word2Vec & Conv-4 & 63.28$\pm$0.95 & 5.31\\ 
         GloVe & Conv-4 & 63.86$\pm$0.91 & \textbf{5.89}\\
         fastText & Conv-4 & 61.77$\pm$0.98 & 3.80\\
         \bottomrule
        \end{tabular}
        }
        \label{table:ablation_different_embeddings}
    }
    \vspace{-1.2em}
\end{table}

\begin{figure}[!htp]
    \RawFloats
   \begin{minipage}{0.6\textwidth}
     \centering
     \includegraphics[width=.9\linewidth]{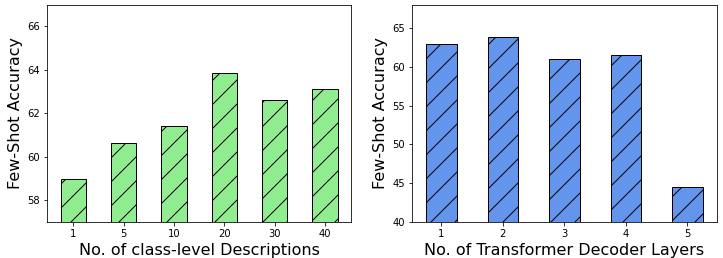}
     \caption{Performance upon varying the number of class-level descriptions (left) and the number of Transformer decoder layers (right).}\label{fig:ablation_varying_descriptions}
   \end{minipage}\hfill
   \begin{minipage}{0.38\textwidth}
     \centering
     \includegraphics[width=.9\linewidth]{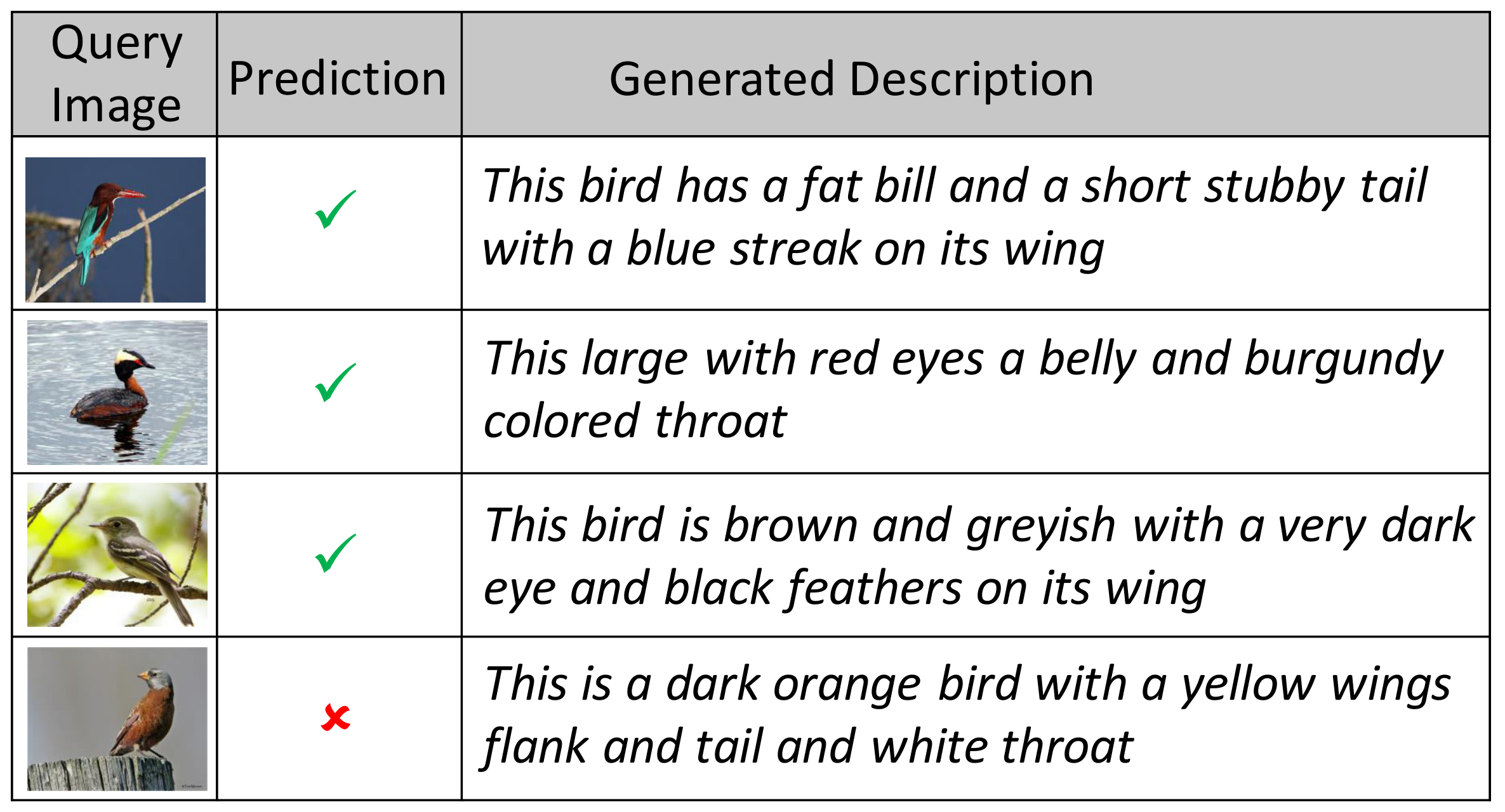}
     \caption{Class-level descriptions generated by RS-FSL for novel query images during inference.}\label{fig:inference_descriptions}
   \end{minipage}
\end{figure}

\noindent \textbf{Number of class-level descriptions.}
Fig.~\ref{fig:ablation_varying_descriptions} (left) shows that upon increasing the number of class-level descriptions from 1 to 20 the accuracy peaks to a maximum of $63.86\%$, however, it starts to saturate after increasing beyond 20. This could be because beyond a certain number of class-level descriptions, the semantic attributes collected from the available descriptions possibly become saturated, rendering the additional descriptions redundant. 


\noindent Fig.~\ref{fig:ablation_varying_descriptions} (right) shows that our method retains the highest accuracy ($63.86\%$) when employing two Transformer layers. However, it starts to deteriorate after further increasing the number of layers i.e. 3 to 5. The inferior performance is most likely due to over-fitting caused by the over-parameterization of the language description model given a relatively small dataset.
%

\begin{SCtable}
\centering\scalebox{0.8}{
\begin{tabular}{l c c c} 
 \toprule
 \rowcolor{mygray} \textbf{Method} & \textbf{Backbone} & \textbf{Accuracy} \\ [0.5ex] 
 \midrule
 Forward Decoder  & Conv-4 & 62.03 ± 0.93  \\ 
 Bidirectional Decoder & Conv-4 & 63.86 ± 0.91 \\
 \bottomrule
\end{tabular}}
\caption{Performance using only the forward decoding and the developed bidirectional decoding.}
\label{table:ablation_forward_only}
\end{SCtable}
\noindent \textbf{Language decoding mechanisms.} 
We replace the bidirectional language decoding mechanism with just the forward decoding and observe that the former improves accuracy by $1.83\%$ compared to latter (Tab.~\ref{table:ablation_forward_only}). Bidirectional decoding can better relate the visual cues with language semantics as it can model two-way interaction between the tokens, thereby facilitating the learning of generalizable visual representations vital for few-shot scenarios.

\noindent \textbf{Auxiliary self-supervision.} We compare the performance of different auxiliary self-supervi-\\sed approaches with our proposed method (Tab.~\ref{table:ablations_different_self_supervisions}). The first auxiliary self-supervised task is predicting the rotation angle of the visual input \cite{rizve2021exploring,rajasegaran2020self}, and the others are predicting language using LSTM-GRU based recurrent architecture (ProtoNet +GRU) \cite{mu2020shaping} and predicting the semantic word embedding corresponding to the prototype (ProtoNet + Word Embeddings).  We observe that the proposed transformer based (bidirectional) language decoding mechanism significantly improves the performance ($5.97\%$) over the method that is not using any auxiliary self-supervision (ProtoNet (without semantics)). Further, our approach outperforms the other auxiliary self-supervision methods, Proto+Rotation, Proto+Word Embeddings and Proto+GRU, by a margin of $4.6\%$, $3.61\%$ and $2.62\%$ respectively.


Fig.~\ref{fig:inference_descriptions} shows class-level descriptions generated by RS-FSL for novel query images (in CUB dataset) during inference. We note that generated descriptions allow us interpreting the model behaviour for incorrect prediction, \eg finding the bird attributes that are confused.

\begin{SCtable}[][h]
\centering\scalebox{0.8}{
\begin{tabular}{l c c} 
 \toprule
 \rowcolor{mygray} \textbf{Method} & \textbf{Accuracy} \\ [0.5ex] 
 \midrule
 ProtoNet (without semantics) & 57.97$\pm$0.96  \\ 
 ProtoNet + Rotation & 59.20$\pm$0.97 \\
 ProtoNet + Word Embeddings & 60.25$\pm$0.93 \\
 ProtoNet + GRU \cite{mu2020shaping} & 61.24$\pm$0.96 \\
 RS-FSL + Class Descriptions & \textbf{63.86$\pm$0.91} \\
 \bottomrule
\end{tabular}}
\caption{Comparison between different auxiliary training methods. Average few-shot 5-way 1-shot accuracy reported with 95\% confidence interval}
\label{table:ablations_different_self_supervisions}
\end{SCtable}

\noindent \textbf{Extra Cost vs Performance boost.} We use Nvidia-RTX Quadro 6000 single-GPU for our training. We observed that training in miniImageNet dataset without auxiliary supervision took around 4 hours for training while RS-FSL took 5 hours with additional transformer decoder layers. Further our model obtains a performance boost of 2\% over the baseline in miniImageNet (Tab. \ref{table:SOTA_cub_mini} left).

\vspace{-1em}
\section{Conclusion}
\vspace{-0.5em}
We presented a new FSL approach that models rich semantics shared across few-shot examples. This is realized by leveraging class-level descriptions, available with less annotation effort. We create a hybrid prototype which is used to produce class-level language predictions as an auxiliary task while training. We develop a Transformer based bi-directional decoding mechanism to connect visual cues with semantic  descriptions to enrich the visual features. Experiments on four datasets show the benefit of our approach. 

\bibliography{egbib}
\newpage

\section*{\textit{Appendix}\\
        \large{Rich Semantics Improve Few-shot Learning}}
        \vspace{0.5em}

\begin{figure}
    \centering
    \includegraphics[width=0.99 \linewidth]{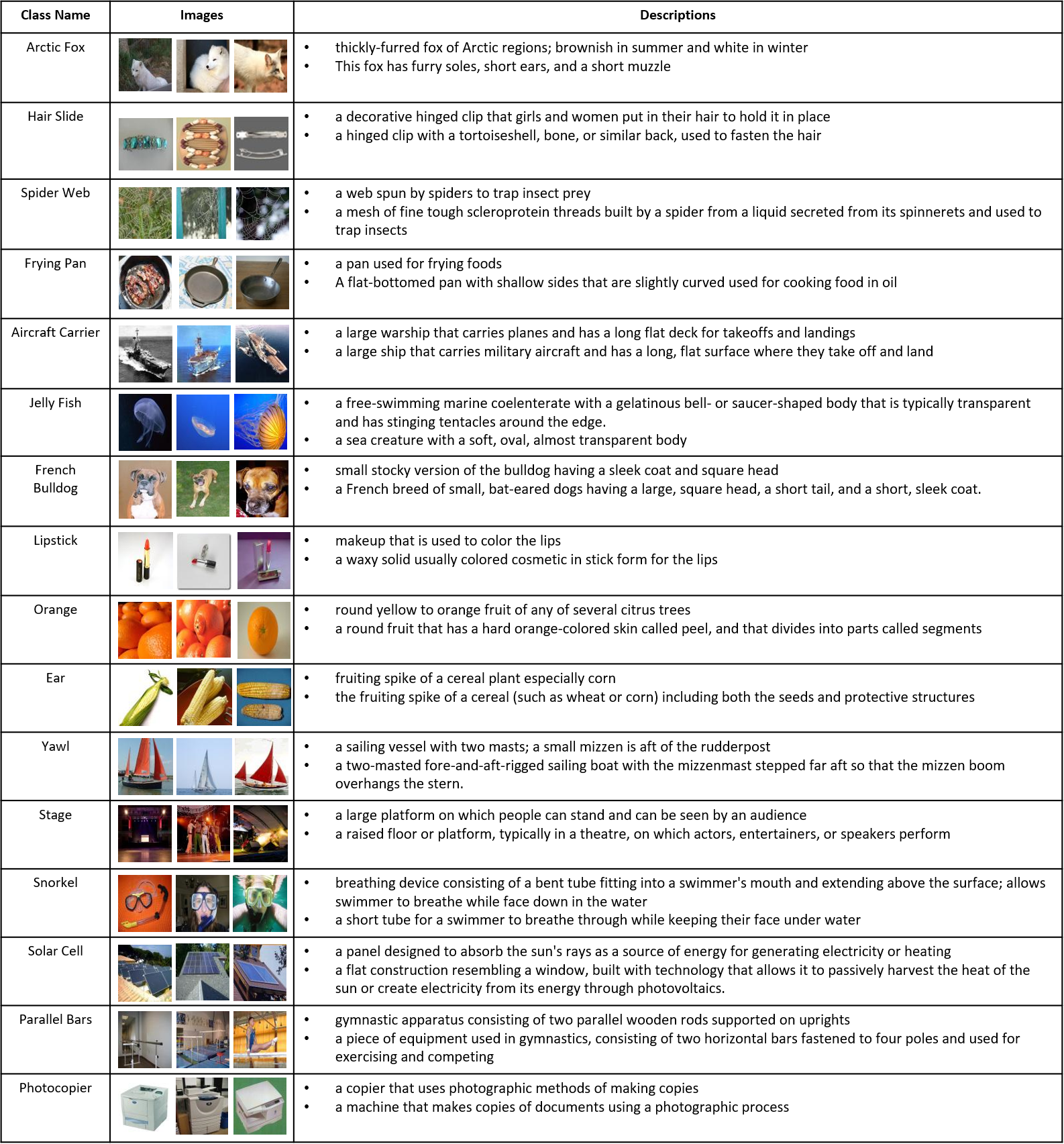}
    \caption{\small Some representative examples of class-level descriptions for miniImageNet dataset. As these descriptions are not readily available, we manually collected them from the web.}
    \label{eg_description}
\end{figure}

\paragraph{Qualitative Examples.}
\noindent Fig.~\ref{eg_description} displays some examples of class-level descriptions for miniImageNet dataset. Since these descriptions were not readily available, we manually collected them from the web. For a given class, all descriptions have been collected such that they mostly represent the shared semantics across that class. For instance, for a frying pan class, the second description has attributes like `flat-bottomed pan', `shallow sides', and `slightly curved' that are shared by the three shown examples belonging to this class. \\
\par
\begin{figure}
    \centering
    \includegraphics[width=0.99 \linewidth]{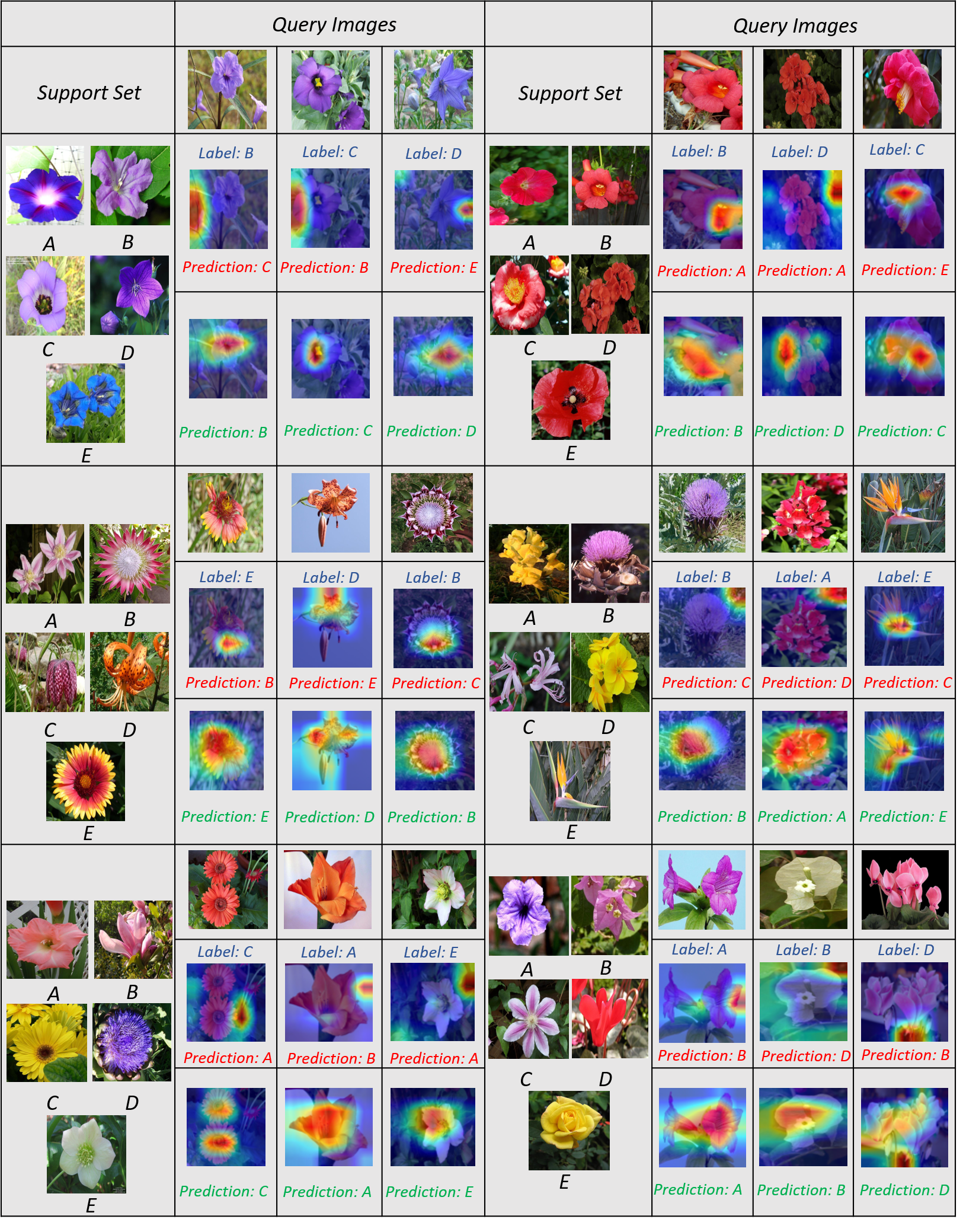}
    \caption{\small Example predictions with corresponding attention maps from the meta-baseline  (red) and our method (green) on five different few-shot tasks in VGG-Flowers dataset. Ground-truth labels are shown in blue. We note improvements in our attention maps leading to correct predictions in visually similar classes.}
    \label{eg_attn_maps}
\end{figure}

\noindent Fig.~\ref{eg_attn_maps} visualizes qualitative examples, where RS-FSL encourages the model to focus on semantically relevant visual regions in order to obtain a correct classification. Specifically, we show example predictions with corresponding attention maps from the meta-baseline \cite{chen2020new} (red) and our method (green) on five different few-shot tasks in VGG-Flowers dataset. Ground-truth labels are shown in blue. We note that, for all query images, RS-FSL produces better attention maps, focused on semantically relevant visual features, which allows it to produce correct classifications. \\

\leavevmode\thispagestyle{empty}\newpage \clearpage
\end{document}